\documentclass[10pt, a4paper, logo]{googledeepmind}

\usepackage{natbib}
\usepackage{float}
\usepackage{algorithm}
\usepackage{algpseudocode}
\usepackage{placeins}
\usepackage[normalem]{ulem}

\usepackage{amsmath,amsfonts,bm}
\usepackage{natbib}
\usepackage{fancyhdr}

\def\eqref#1{equation~\ref{#1}}
\def\Eqref#1{Equation~\ref{#1}}

\def\1{\bm{1}}

\def\vc{{\bm{c}}}

\def\vx{{\bm{x}}}
\def\vy{{\bm{y}}}

\def\vtau{{\bm{\tau}}}

\DeclareMathAlphabet{\mathsfit}{\encodingdefault}{\sfdefault}{m}{sl}
\SetMathAlphabet{\mathsfit}{bold}{\encodingdefault}{\sfdefault}{bx}{n}

\def\gA{{\mathcal{A}}}

\def\gD{{\mathcal{D}}}

\def\gM{{\mathcal{M}}}

\def\gS{{\mathcal{S}}}

\def\gU{{\mathcal{U}}}

\def\bbE{{\mathbb{E}}}

\newcolumntype{Y}{>{\raggedright\arraybackslash}X}
\newcommand{\method}{Skill-RM}
\definecolor{promptpurple}{RGB}{242, 242, 255}
\definecolor{promptgreen}{RGB}{242, 255, 242}
\definecolor{promptblue}{RGB}{242, 248, 255}
\definecolor{skillback}{RGB}{255, 252, 242}
\definecolor{bordergray}{RGB}{80, 80, 80}
\newtcolorbox{promptbox}[1]{
  breakable,
  enhanced,
  colback=#1,
  colframe=bordergray,
  boxrule=1pt,
  sharp corners,
  left=15pt,
  right=15pt,
  top=12pt,
  bottom=12pt,
  before skip=6pt,
  after skip=10pt,
  fontupper=\ttfamily
}
\newtcolorbox{skillbox}{
  breakable,
  enhanced,
  colback=skillback,
  colframe=bordergray,
  boxrule=1pt,
  sharp corners,
  left=15pt,
  right=15pt,
  top=12pt,
  bottom=12pt,
  before skip=6pt,
  after skip=10pt,
  fontupper=\small\ttfamily
}

\newcommand{\skill}{\mathcal{S}}

\newcommand{\Ylist}{\bm{Y}}

\setlist[itemize]{leftmargin=*, itemsep=2pt, topsep=3pt}
\setlist[enumerate]{leftmargin=*, itemsep=2pt, topsep=3pt}

\title{Skill-RM: Unifying Heterogeneous Evaluation Criteria via Agent Skill}

\author[1,2]{Tao Chen\textsuperscript{\dag}}
\author[1]{Gangwei Jiang}
\author[1]{Pengyu~Cheng\textsuperscript{\S}}
\author[1,3]{Siyuan Huang\textsuperscript{\dag}}
\author[1,4]{Yihao Liu\textsuperscript{\dag}}
\author[1,5,6]{Jingwei~Ni\textsuperscript{\dag}}
\author[1]{Jiaqi~Guo}
\author[1]{Mengyu~Zhou}
\author[1]{Kai Tang}
\author[1]{Junling Liu}
\author[2]{Qinliang Su\textsuperscript{\S}}
\author[1]{Xiaoxi Jiang}
\author[1]{Guanjun Jiang}
\affil[1]{Qwen Large Model Application Team, Alibaba}
\affil[2]{Sun Yat-sen University}
\affil[3]{The Chinese University of Hong Kong}
\affil[4]{Peking~University}
\affil[5]{ETH~Z\"urich}
\affil[6]{University of Zurich}

\begin{abstract}

Reward models (RMs) provide critical feedback signals for LLM post-training, notably in reinforced fine-tuning (RFT) and reinforcement learning (RL) pipelines. However, current reward evaluation rely on heterogenous criteria such as rule-based verifiers, ground-truth references, procedural checklists, and complex rubrics, where a unified mechanism to integrate all types of evidences remains unexplored.
To this end, we propose Skill Reward Model (Skill-RM), a unified framework that reformulates reward modeling as the execution of a reusable Reward-Evaluation Skill. By treating reward computation as a structured agentic task, Skill-RM provides a consistent interface to orchestrate heterogeneous resources, dynamically selecting and aggregating evidence tailored to the specific requirements of each input. This approach enables the reward model to move beyond static evaluation, ensuring consistency and transparency across diverse tasks. Extensive experiments on reward benchmarks and downstream applications, including best-of-N selection and reinforcement learning, demonstrate that Skill-RM consistently outperforms traditional judge baselines. Our findings suggest that Skill-RM not only provides a unified solution for reward modeling but also achieves superior performance through the strategic and dynamic orchestration of evidence. The code is at {\small \url{https://github.com/Qwen-Applications/Skill-RM}}.
\end{abstract}

\begin{document}
\maketitle

\begin{figure}[h]
\centering
\includegraphics[width=0.97\textwidth]{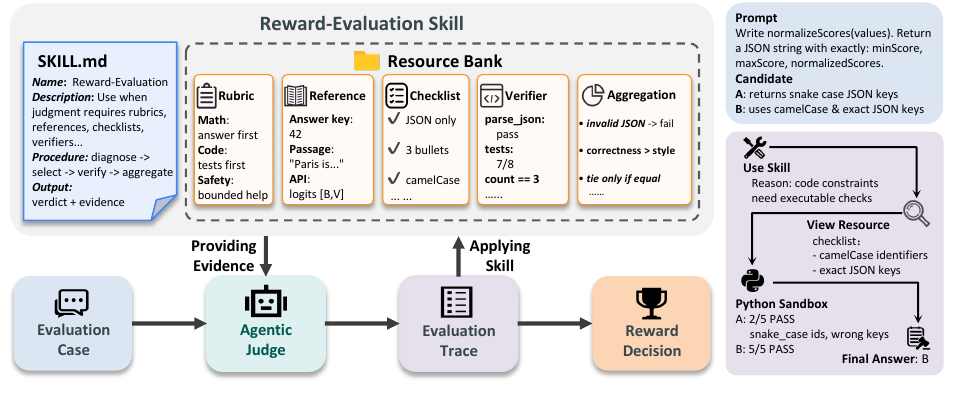}
\vspace{-2mm}
\caption{ 
 Overview of Skill Reward Model (Skill-RM). The Reward-Evaluation Skill comprises a procedural document and a structured resource bank (including rubrics, checklists, verifiers, and aggregation rules). During evaluation, Skill-RM dynamically retrieves relevant resources and executes an agentic evaluation trace.  This input-adaptive reasoning process effectively unifies diverse reward evaluation paradigms into a consistent and adaptable framework.
}
\label{fig:skillrm_framework}
\end{figure}

\vspace{-2mm}
\section{Introduction}
\vspace{-2mm}


Reward models (RMs) are crucial in the post-training of large language models (LLMs)~\citep{bai2022training,cheng2023everyone,zeng2024diversified,genrm,agenticrewardmodeling}, for their primary supervisory mechanism to steer models toward desired behaviors during reinforcement learning (RL)~\citep{ouyang2022training} and reinforced fine-tuning (RFT)~\citep{cheng2024adversarial,reft}. Traditional usage of RMs in LLM post-training is straightforward, including re-ranking candidate responses for best-of-N selection~\citep{llama2}, curating high-quality training data~\citep{wizardlm,skyworkrewardv2}, acting as automated judges~\citep{zheng2023judging}, and generating feedback signals for downstream alignment~\citep{rafailov2023direct,zheng2023judging,rlaif,rewardbench2}.
%

The rapid advancement of LLM capabilities in reasoning~\citep{wei2022chain}, coding~\citep{codex}, and tool-use~\citep{toollm} raises increasingly sophisticated demands on reward modeling~\citep{agenticrewardmodeling,rewardbench2,tirjudge}. Consequently, the RM evaluation criteria to supervise these models evolve from simple scalar preferences into complex, resource-intensive verification processes.
%
%
%
 For instance, ensuring factual accuracy now necessitates external references or retrieved evidence~\citep{nakano2021webgpt,factscore}. Mathematical and coding tasks demand verifiable execution against ground truths~\citep{lightman2024let}. Agentic training requires validating multi-step tool trajectories with additional tool calling~\citep{toollm,lu2025search,tirjudge}. Safety enforcement often hinges on strict constraint decomposition or policy-based vetoes~\citep{bai2022constitutional,jiang2024followbench}. These tasks demonstrate that high-fidelity reward assignment is no longer a monolithic task but a multi-faceted process involving heterogeneous resources. 



However, existing RM designs struggle to integrate such diversified resource-dependent judgments into a coherent framework. Scalar RMs compress complex, resource-grounded evidence into opaque scores, rendering the evaluation process fundamentally uninterpretable and inflexible~\citep{armorm}. LLM-as-a-Judge systems offer natural language reasoning capabilities~\citep{liu2023g, zheng2023judging}. They typically rely on unstructured, flat-prompting, where rubrics, examples, and tools are concatenated into a single prompt. This approach leaves critical aspects (such as resource selection, evidence tracking, and signal aggregation) implicit and unmanaged. While recent efforts introduce criteria-conditioned, rubric-centered, or tool-augmented evaluation systems~\citep{prometheus2,verif,agenticrewardmodeling,openrs,tirjudge}, they primarily focus on isolated mechanisms, exposing only one resource modality at a time. Consequently, there is a critical need for a unified, reusable abstraction that can seamlessly synthesize heterogeneous resources, adaptively select evidence based on the input, and facilitate explicit, evidence-grounded reward computation across diverse evaluation scenarios.

On the other hand, agent skills provide a natural way to instantiate this missing abstraction. In recent agentic research, an agent skill is defined as a self-contained, filesystem-based package of procedural logic and supporting resources that an LLM agent can discover, load, and execute on demand~\citep{claude2025introducingskills,xu2026agentskills}. 
The standard implementation of an agent skill contains a main document (\texttt{SKILL.md}) to encode the agentic procedure, auxiliary scripts and assets available for on-demand loading, and metadata to facilitate discovery and versioning~\citep{zhang2025equippingagents,agentskills2025standard}.
%
%
This design fundamentally transcends both flat prompting (which merely adds context) and atomic tool-use (which exposes isolated executables); instead, a skill packages the orchestration logic that determines \textit{what} resources to consult, \textit{when} to invoke them, and \textit{how} their outputs should synthesize the agent's final decision~\citep{claude2025skillsexplained,xu2026agentskills,jiang2026agenticskills}. Coherently, we can map this principle directly onto heterogeneous evaluation criteria: rubrics, references, constraints, and verifiers are modular and loadable resources, while a "reward skill" governs their invocation and synthesis. 

To this end, we propose \textbf{Skill Reward Model} (Skill-RM), a unified framework that reformulates reward modeling as the execution of a reusable \textit{Reward-Evaluation Skill}. This skill encapsulates evaluation logic through an explicit invocation protocol, interfaces to heterogeneous resources, and a structured schema for evidence collection. As in Figure~\ref{fig:skillrm_framework}, when given an evaluation instance, Skill-RM executes the skill via a systematic workflow: it identifies applicable criteria, selectively invokes relevant resources, and aggregates criterion-level evidence into a final reward. 
%
Skill-RM reframes reward modeling from the passive absorption of evaluation knowledge (either within model parameters or unstructured prompt context) to the active orchestration of the selection and execution of resources.
Consequently, reward computation becomes more adaptable and interpretable across heterogeneous tasks.
%
%
Extensive experiments across major benchmarks, including best-of-N selection and RFT, demonstrate that Skill-RM consistently outperforms traditional judge baselines. Our ablation studies further confirm that these gains stem from the skill-mediated orchestration of evidence rather than merely increasing context or tool availability. Together, these results validate that structuring evaluation as an executable procedure significantly enhances reward quality, providing a scalable path toward higher-quality LLM post-training and judgment. 
%

\vspace{-1.5mm}
\section{Preliminary}
\label{sec:prelim}
\vspace{-1.5mm}

\paragraph{Pointwise Reward Modeling}
Denote \(\vx\in\mathcal{X}\) as a user prompt, and \(\vy\in\mathcal{Y}\) as the corresponding response generated by an LLM policy $\pi_{\theta}(\vy|\vx)$. 
A pointwise reward model $r_\phi(\cdot)$ assigns a scalar score to a single response, denoted as $r_\phi(\vx, \vy)$. In traditional policy optimization, such as RLHF~\citep{ouyang2022training}, the policy $\pi_\theta(\vy|\vx)$ is optimized by maximizing the expected reward:
\begin{equation} \label{eq:obj-rlhf}
    \max_{\pi_\theta} \mathbb{E}_{\vx \sim \mathcal{D}, \vy \sim \pi_\theta(\vy|\vx)} [r_\phi(\vx, \vy)],
\end{equation}
where $\mathcal{D}$ denotes the distribution of prompts, and $\vy$ is sampled from the learning LLM policy $\pi_\theta(\vy|\vx)$. To solve this optimization problem for LLM post-training, existing paradigms include direct preference optimization (DPO)~\citep{rafailov2023direct}, which implicitly recovers the reward function, as well as recent reinforcement learning approaches such as GRPO~\citep{deepseekmath} and GSPO~\citep{gspo} 
, which leverage point-wise RMs to provide scalable feedback signals.

\paragraph{Pairwise Reward Modeling}
A pairwise reward model directly compares two responses and provides a binary output $\hat{r}_\phi(\vx,\vy,\bar{\vy})$ indicating whether $\vy$ is preferred to $\bar{\vy}$, where $\hat{r}_\phi(\vx,\vy,\bar{\vy}) = 1$ denotes $\vy \succ \bar{\vy}$ (\textit{i.e.}, $\vy$ is preferred over $\bar{\vy}$), and $\hat{r}_\phi(\vx,\vy,\bar{\vy}) = 0$ otherwise. Mathematically, the output can be interpreted as a sample from a Bernoulli preference distribution with probability $\mathbb{P}_\phi[\vy \succ \bar{\vy} | \vx]$~\citep{rafailov2023direct}.
%
%
A prominent implementation of this paradigm is the Generative Reward Model (GRM), which frames preference estimation as a conditional text generation task. Unlike scalar-based models that output a single opaque value, GRMs produce structured rationales or critiques before predicting the final preference label~\citep{genrm,deepseekgrm,thinkrm}. This generative approach is increasingly adopted for nuanced, subjective judgments, such as the self-critique mechanisms employed in state-of-the-art models like Kimi-k2~\citep{kimik2}. Furthermore, these pairwise preferences provide a robust optimization signal for preference-based reinforcement learning, as exemplified by algorithms such as IPO~\citep{azar2023general}:
\begin{equation}
    \max_{\pi_\theta}
   \bbE_{
   \vx \sim \gD,
   \vy \sim \pi_\theta(\cdot|\vx),
        \bar{\vy} \sim \mu(\cdot|\vx)
        }
    \left[
        \mathbb{P}_\phi[
        \vy \succ \bar{\vy}\mid\vx]
    \right],
    \label{eq:pairwise-rm-rl}
\end{equation}
where \(\mu(\cdot|\vx)\) is a reference policy for pairwise comparison.

\paragraph{Rubric-conditioned Reward Modeling}
Rubric-conditioned reward modeling makes the evaluation criteria explicit instead of leaving them fully implicit in model parameters. Let \(\vc=(c_1,c_2,\ldots,c_M)\) be a set of rubrics or checklists. A rubric-conditioned reward model evaluates \((\vx,\vy)\) with respect to these criteria and first produces criterion-level judgments \(s_m=r_\phi(\vx,\vy;c_m)\). The final reward is then obtained by an aggregation rule
\begin{equation}
    r_\phi(\vx,\vy;\vc)
    =
    \gA(s_1,\ldots,s_M),
    \label{eq:rubric-rm}
\end{equation}
where \(\gA\) can combine criterion-level judgments from rubric-based judges~\citep{prometheus2023} through checklist-based scoring~\citep{rlcf}, criterion-wise rubric aggregation~\citep{openrs}, or calibrated aggregation over rubric-question outputs~\citep{llmrubric}. 

\paragraph{Agent Skills}
An agent skill packages reusable procedural knowledge for a specific type of tasks as a loadable artifact. It exposes lightweight metadata for discovery, loads its main instructions only when relevant, and consults auxiliary files or executable resources on demand~\citep{claude2025introducingskills,zhang2025equippingagents,agentskills2025standard,xu2026agentskills}. Formally, this file-based design is defined as:
\begin{equation}
    \skill=(\gM,\gU),
    \quad
    \gU=\{u_1,u_2,\dots,u_K\},
    \label{eq:skill-definition}
\end{equation}
where $\gM$ denotes the skill specification (\texttt{SKILL.md}) including routing metadata and main procedural instructions. The set \(\gU\) denotes an auxiliary resource bank co-located with the skill, which includes references, tools, checklists, verifiers, \textit{etc}. This separation supports progressive disclosure, that irrelevant resources need not enter the context, while relevant resources can be inspected or executed during skill-use procedure~\citep{ling2026agentskillsanalysis,jiang2026agenticskills}. 

\vspace{-1.5mm}
\section{Skill Reward Model}\label{sec:method}
\vspace{-1.5mm}

We propose \textbf{Skill Reward Model} (Skill-RM), a unified framework that reformulates reward modeling from a monolithic scoring task into a \textbf{skill-mediated execution procedure}. Rather than eliciting rewards through opaque parameter-based scoring or unstructured, flat-prompting, \method{} treats reward computation as the systematic execution of a reusable \textit{Reward-Evaluation Skill}. This skill governs the entire evaluation lifecycle—from resource orchestration and criterion-level evidence synthesis to final reward aggregation—thereby rendering reward computation instance-adaptive, evidence-grounded, and modular.


As in Figure~\ref{fig:skillrm_framework}, Skill-RM comprises three core components: (1) a \emph{Reward-Evaluation Skill} that defines the invocation protocol and manages a resource bank of heterogeneous evaluation resources; (2) a \emph{skill-mediated evaluation process}, where an agentic judge incrementally retrieves task-relevant resources to derive structured, criterion-level evidence; and (3) a deterministic \emph{reward readout function} that maps this structured trace to the task-required reward output. The fundamental novelty of \method{} lies in shifting the paradigm: instead of merely providing additional evaluation resources, we formalize their selection, execution, and synthesis into an explicit, reproducible computation procedure.

\vspace{-1.5mm}
\subsection{Reward-Evaluation Skill}
\vspace{-1mm}

We instantiate reward evaluation as a reusable \textit{Reward-Evaluation Skill} defined as:
\begin{equation}
    \mathcal{S}_{\text{RM}} = (\mathcal{M}_{\text{RM}}, \mathcal{U}_{\text{RM}}),
    \label{eq:rm-skill}
\end{equation}
where $\mathcal{M}_{\text{RM}}$ denotes the \textbf{procedural skill specification} and $\mathcal{U}_{\text{RM}}$ represents the \textbf{resource bank}. The specification $\mathcal{M}_{\text{RM}}$ explicitly governs: (i) the definition and scope of evaluation criteria;
    (ii) the invocation protocol for resource selection and execution;
    (iii) the schema for systematic evidence collection; and
    (iv) the output contract for the resulting structured judgment $z$.
Complementing this, $\mathcal{U}_{\text{RM}}$ houses the concrete materials and executable interfaces necessary for task-specific verification.


A fundamental design philosophy of Skill-RM is that reward knowledge is not implicitly buried within model weights or compressed into a monolithic prompt. Instead, it is \textit{externalized} as a reusable skill across diverse instances and domains. This skill serves as the atomic unit of reward capability, encoding a comprehensive evaluation logic: it dictates not only \textit{what} criteria should be assessed, but also \textit{how} evidence is retrieved and \textit{how} it is synthesized to substantiate the final reward decision.

To support criterion-level auditability, the skill constrains the final judgment to be evidence-bearing rather than purely decision-bearing. 
 Let $\mathcal{C} = \{c_1, \ldots, c_M\}$ denote the set of evaluation criteria activated by $\mathcal{M}_{\text{RM}}$. For each criterion $c_m$, the judge model generates a local evidence item:
 \begin{equation}
e_m=(c_m, q_m, s_m),
\label{eq:criterion-evidence}
\end{equation}
where $q_m$ denotes the supporting observation collected from invoked resources,  
and $s_m$ signifies the local assessment—such as binary states (e.g., satisfied/violated), uncertainty, or criterion-specific scores. The final structured judgment $z$ is then defined as:
\begin{equation}
    z = (\mathcal{E}, d), \quad \mathcal{E} = \{e_m\}_{m=1}^{M},
    \label{eq:structured-judgment}
\end{equation}
where $d$ acts as the conclusive decision field from which the reward output is deterministically read out. 
This design ensures that the ultimate reward output is not an opaque scalar, but is explicitly traceable to the underlying evidence $\mathcal{E}$ for each criterion.


\vspace{-1.5mm}
\subsection{Reward-Evaluation Resources}
\vspace{-1mm}
\begin{table*}[t]
\centering
\small
\setlength{\tabcolsep}{5pt}
\renewcommand{\arraystretch}{1.08}
\begin{tabularx}{0.98\textwidth}{@{}p{0.20\textwidth}p{0.30\textwidth}Y@{}}
\toprule
\textbf{Resource Type} & \textbf{Examples} & \textbf{Role in judgment} \\
\midrule
Rubric \& Criterion
& Helpfulness, correctness, safety
& Defines judgment dimensions and relative priorities. \\
\addlinespace[4pt]
Reference
& Answer key, evidence passage, official solution
& Grounds factual, mathematical, or task-specific correctness checks. \\
\addlinespace[4pt]
Checklist \& Constraint
& Required response format, coverage requirement, forbidden behavior
& Decomposes instruction following into checkable conditions. \\
\addlinespace[4pt]
Verifier \& Tool
& Python sandbox, code checker, exact-match checker
& Produces executable observations from visible inputs or available references. \\
\addlinespace[4pt]
Calibration \& Aggregation rule
& Score calibration, bias control, evidence priority
& Resolves evidence conflicts and maps observations to a judgment. \\
\bottomrule
\end{tabularx}
\vspace{-2mm}
\caption{Resource types in the Reward-Evaluation Skill. Each type contributes either evaluation criteria, evidence-producing procedures, or judgment calibration for skill-mediated reward evaluation.}
\label{tab:evaluation-resources}
\vspace{-1.5mm}
\end{table*}

We define a reward-evaluation resource as a structured unit of task-relevant evidence or executable evaluation procedure that can be consulted during skill execution. Formally, let
\begin{equation}
    \gU_{\text{RM}} = \{ u_1, u_2, \ldots, u_K \}
    \label{eq:resource-set}
\end{equation}
denote the resource bank exposed by the Reward-Evaluation Skill, where each resource
\begin{equation}
    u_i = (\texttt{type}_i, \texttt{content}_i, \texttt{scope}_i, \texttt{act}_i)
    \label{eq:resource-definition}
\end{equation}
is specified by its type, semantic content, applicability scope, and access action. The access action \(\texttt{act}_i\) can be inspecting textual content (\textit{e.g.}, a rubric or reference), executing a deterministic procedure (\textit{e.g.}, a verifier or tool), or retrieving a rule for calibration and aggregation. 
%
%
%
The resource bank $\mathcal{U}_{\text{RM}}$ encompasses a diverse spectrum of evaluation tools spanning the full reward computation lifecycle: analyzing evaluation criteria, obtaining verifiable observations, and aggregating evidence into a final judgment. Table~\ref{tab:evaluation-resources} provides a taxonomy of resources and their functional roles in the judgment.
Critically, unlike traditional methods that overwhelm the judge with monolithic and flat contexts, our framework maintains a latent resource repository. Resources remain inactive until triggered by the skill specification $\mathcal{M}_{\text{RM}}$, which ensures that only the task-relevant subset of resources is disclosed and invoked during the judgment process, thereby minimizing contextual noise and maximizing the precision of the agentic judge.


We construct the resource bank through an LLM-assisted curation pipeline designed to ensure consistency and modularity. We aggregate candidate resources from a diverse spectrum of sources, including existing reward-modeling literature, standardized judge protocols, benchmark documentation, and verifiable evaluation practices. To ensure reusability, we define explicit applicability conditions for each resource, merge redundant entries, and strip away task-specific heuristics to produce generalized modules. The finalized resource bank is then version-controlled and frozen before evaluation to ensure reproducibility. Further details regarding the construction and cataloging of these resources are provided in Appendix~\ref{app:skill-construction}.

\vspace{-1.5mm}
\subsection{Skill-Mediated Judgment}
\vspace{-1mm}

Given an input prompt $\vx$ and a set of candidate responses $\mathcal{Y} = \{\vy_1, \dots, \vy_K\}$, Skill-RM performs reward evaluation by invoking the skill artifact $\mathcal{S}_{\text{RM}}$ via an agentic judge model $\pi_\phi$. Upon initiation, the judge loads the procedural specification $\mathcal{M}_{\text{RM}}$, which serves as the execution blueprint for the current evaluation task. Throughout the process, the judge dynamically retrieves or executes the resources in $\mathcal{U}_{\text{RM}}$ on demand, strictly adhering to the invocation protocol defined by $\mathcal{M}_{\text{RM}}$. This ensures that the evaluation is not a single-pass inference, but a sequence of orchestrated actions tailored to the specific requirements of the input.
Formally, the evaluation process is an action-observation trace:
\begin{equation}
    \vtau
    =
    (a_1,o_1,\ldots,a_T,o_T,z) \sim
    \pi_\phi(\cdot \mid \vx,\Ylist;\gS_\text{RM}),\quad \mathcal{S}_{\text{RM}} = (\mathcal{M}_{\text{RM}}, \mathcal{U}_{\text{RM}}), 
    \label{eq:skill-judgment-trace}
\end{equation}
where \(a_t\) is a judge action at $t$-th step, \(o_t\) is the corresponding returned observation, and \(z\) is the final structured judgment defined in \Eqref{eq:structured-judgment}. The action set $\{a_t\}$ contains listing available resources, inspecting evidence, executing tools, and finalizing the judgment.
%

Operationally, the evaluation follows a staged execution logic: (1) identifying the evaluation targets and the desired output format; (2) activating the relevant criteria specified by $\mathcal{M}_{\text{RM}}$; (3) dynamically retrieving the necessary resources. As the process unfolds, the judge maps these observations into criterion-level evidence, ensuring that the structured judgment $z$ is populated only after all mandatory evidence fields are satisfied. By formalizing this process, reward evaluation is transformed into a rigorous, protocol-governed computation, replacing the heuristic-based, one-shot decisions characteristic of traditional prompt-level reward modeling.

\vspace{-1.5mm}
\subsection{Reward Readout}
\vspace{-1mm}
The task-required reward output is derived via a deterministic readout function $\gA(\cdot)$: 
\begin{equation}
    r^\text{Skill}_\phi(\vx,\Ylist;\gS_\text{RM})
    =
    \gA(\vtau),\quad \mathcal{S}_{\text{RM}} = (\mathcal{M}_{\text{RM}}, \mathcal{U}_{\text{RM}}),
    \label{eq:skill-readout}
\end{equation}
where the skill specification \(\gM_{\text{RM}}\) constrains the form of the structured judgment \(z\). The readout function $\mathcal{A}(\cdot)$ parses $z$ from the execution trace $\tau$ to project it into the task-specific reward-output space:
\begin{equation}
\gA(\vtau)\in
\begin{cases}
\mathbb{R}, & \text{if } K=1 \text{ (pointwise)}, \\
\{1, \ldots, K\}, & \text{if } K\ge 2 \text{ (selection)}.
\end{cases}
\label{eq:skill-readout-space}
\end{equation}
%
Pairwise comparison is treated as a specialized case ($K=2$), where the readout function interprets the judgment to yield a preference decision. 
Consequently, pointwise scoring and multi-candidate selection are unified as two distinct projections of the same underlying evidence-bearing process. 


\vspace{-1.5mm}
\section{Related Work}
\vspace{-1.5mm}

\paragraph{Reward Modeling \& Reasoning.}
Traditional reward modeling for language-model alignment is framed as scalar preference prediction~\citep{bai2022training,cheng2023everyone,Li2025EliminatingIB}. InstructGPT~\citep{ouyang2022training} establishes RLHF with pairwise human preferences, while DPO~\citep{rafailov2023direct} shows that the same supervision can train policies without an explicit online reward model. \citet{lambert2024rewardbench,rewardbench2} make reward-model quality a shared evaluation target. ~\citet{armorm,skyworkrewardv2} improve objective decomposition and preference-data scaling. Generative and deliberative reward models further use rationales or self-generated reasoning traces~\citep{genrm}, critiques~\citep{criticrm}, explicit reward-model reasoning~\citep{rmr1}, long-horizon reward reasoning~\citep{thinkrm}, and inference-time scaling with adaptive principles and voting~\citep{deepseekgrm}. 

\paragraph{Resource-aware Evaluation.}
LLM-as-a-Judge systems extend reward-like evaluation by prompting language models to compare, critique, and explain outputs. MT-Bench and Chatbot Arena popularize scalable open-ended assistant evaluation~\citep{zheng2023judging}. Later work analyzes evaluator biases~\citep{cobbler} and decomposes evaluation into fine-grained evaluation dimensions~\citep{flask}. Resource-aware evaluators make the evaluation standard explicit through rubrics, criteria, and references~\citep{prometheus2023,prometheus2}, or through natural-language unit tests~\citep{lmunit}. Other methods expose standards as constitutional principles~\citep{bai2022constitutional}, dynamically provided natural-language reward principles~\citep{rewardanything}, checklist feedback~\citep{rlcf}, synthetic rubrics~\citep{openrubrics}, or pairwise adaptive rubrics with criterion-wise aggregation~\citep{openrs}. Verifiable and tool-augmented evaluation further checks verifiable instruction-following constraints~\citep{ifeval,ifbench}, factual precision against reliable sources~\citep{factscore}, rule- or LLM-based verification signals~\citep{verif,agenticrewardmodeling}, external validation tools~\citep{findeis2025external}, and Python-executor or tool-integrated judge reasoning~\citep{tirjudge}. 

\paragraph{Agent Skills \&  Skill-mediated Methods}
Agent skills have recently emerged as a mechanism for packaging procedural knowledge and task resources into reusable artifacts for LLM agents. Anthropic introduces Agent Skills in Claude as folders of instructions, scripts, and resources~\citep{claude2025introducingskills}, and later describes the design as a way to equip real-world agents with portable procedural knowledge~\citep{zhang2025equippingagents}. The open Agent Skills specification formalizes this design around a \texttt{SKILL.md}-centered directory with optional scripts, references, and assets available through progressive disclosure~\citep{agentskills2025standard}. Earlier skill-library agents such as Voyager show that executable behaviors can be stored and reused across tasks and environments~\citep{wang2023voyager}. Claude's product-level comparison separates Skills from prompts, Projects, MCP, and subagents~\citep{claude2025skillsexplained}. Recent surveys organize agent skills across architecture, acquisition, security, applications, and lifecycle stages~\citep{xu2026agentskills,zhou2026agentskillssurvey}, while systematization work distinguishes agentic skills from atomic tool calls through applicability conditions, execution policies, termination criteria, and reusable interfaces~\citep{jiang2026agenticskills}. Empirical analyses study public Claude skill libraries~\citep{ling2026agentskillsanalysis}, and SkillsBench benchmarks how skill content and curation affect task performance~\citep{li2026skillsbench}. Other work studies structured skill representations~\citep{liang2026skillstructure} and ecosystem-scale skill selection and orchestration~\citep{li2026agentskillos}. \method{} instead specializes the skill abstraction for reward modeling, where the skill organizes reward criteria, references, verifiers, evidence fields, and aggregation rules into an executable evaluation procedure.

\vspace{-1.5mm}
\section{Experiments}
\label{sec:experiments}
\vspace{-1.5mm}

We evaluate \method{} across broad reward-model benchmarks, controlled same-backbone ablations, Best-of-$N$ selection, and instruction-following reward use.
Accordingly, we organize the section around five research questions (RQs). RQ1 asks whether \method{} improves reward benchmarking over matched LLM-as-a-Judge baselines and representative reward-model baselines under the default benchmark protocol. RQ2 asks whether \method{} further benefits when sample-specific resources are mounted through the Reward-Evaluation Skill. RQ3 tests whether the gain comes from skill-mediated resource use rather than from simply adding resources or tools to the prompt. RQ4 evaluates whether \method{} improves fixed-pool best-of-$N$ response selection. RQ5 examines whether \method{} can serve as a reward source for downstream instruction-following RL.

\vspace{-1mm}
\subsection{Experimental Setup}
\vspace{-1mm}

\paragraph{Benchmarks and metrics.} We evaluate three experimental tracks. General reward benchmarking uses RewardBench2~\citep{rewardbench2} for multi-dimensional reward capability, RM-Bench~\citep{rmbench} for content subtlety and style-bias robustness, and JudgeBench~\citep{judgebench} for correctness-focused evaluation; we report each benchmark's default score. Best-of-$N$ selection uses JETTS candidate pools~\citep{jetts}. Instruction-following reward use reports IF-RewardBench Kendall correlation~\citep{ifrewardbench} and downstream IFEval, IFBench, and AdvancedIF scores~\citep{ifeval,ifbench,advancedif}. Unless noted otherwise, ``Avg.'' is the arithmetic mean over reported complete columns.

\begin{table}[t]
\centering
\small
\setlength{\tabcolsep}{5.2pt}
\renewcommand{\arraystretch}{1.10}
\begin{tabular}{@{}lcccc@{}}
\toprule
\textbf{Method} & \textbf{RewardBench2} & \textbf{RM-Bench} & \textbf{JudgeBench} & \textbf{Avg.} \\
\midrule

\multicolumn{5}{@{}l}{\emph{Scalar / learned reward models}} \\
\addlinespace[1pt]
INF-ORM-Llama3.1-70B 
& 76.5 & 75.4 & 70.2 & 74.0 \\
Skywork-Reward-V2-Qwen3-8B 
& 78.2 & 82.6 & 73.4 & 78.1 \\
Skywork-Reward-V2-Llama-3.1-8B 
& \underline{84.1} & \textbf{92.8} & 80.0 & \underline{85.6} \\
\midrule

\multicolumn{5}{@{}l}{\emph{Generative / reasoning reward models}} \\
\addlinespace[1pt]
RM-R1-DeepSeek-Distill-Qwen-32B 
& 71.0\textsuperscript{*} & 83.9 & 56.1\textsuperscript{*} & 70.4 \\
RRM-32B 
& 70.1\textsuperscript{*} & 85.4 & 76.0 & 77.2 \\
RationaleRM-Qwen3-30B-A3B 
& -- & 87.1 & 82.0 & -- \\
\midrule

\multicolumn{5}{@{}l}{\emph{Rubric / resource-based systems}} \\
\addlinespace[1pt]
Auto-Rubric-Qwen3-32B (HelpSteer3 rubrics) 
& 82.3 & 88.1 & 80.9 & 83.8 \\
OpenRubrics Rubric-RM-8B-voting@5 (Qwen3-8B) 
& 67.3\textsuperscript{*} & 80.6\textsuperscript{*} & 44.4\textsuperscript{*} & 64.1 \\
\midrule

\multicolumn{5}{@{}l}{\emph{Agentic / verifier-based judges}} \\
\addlinespace[1pt]
RewardAgent (Qwen3.5-27B) 
& 82.0\textsuperscript{*} & 80.5\textsuperscript{*} & 66.3\textsuperscript{*} & 76.3 \\
TIR-Judge-Zero (Qwen3-8B) 
& 73.4\textsuperscript{\textdagger} & 83.7 & 72.0 & 76.4\textsuperscript{\textdagger} \\
\midrule

\multicolumn{5}{@{}l}{\emph{LLM-as-a-Judge}} \\
\addlinespace[1pt]
GPT-4o Judge 
& 64.9 & 73.1 & 59.8 & 65.9 \\
Claude-3.5-Sonnet Judge 
& 64.7 & 74.5 & 64.8 & 68.0 \\
Qwen3.5-27B 
& 81.1 & 89.8 & 80.8 & 83.9 \\
Qwen3.5-122B-A10B 
& 79.0 & 82.2 & 67.1 & 76.1 \\
\midrule

\multicolumn{5}{@{}l}{\emph{Ours}} \\
\addlinespace[1pt]
\method{} (Qwen3.5-27B) 
& \textbf{85.0} & \underline{91.5} & \underline{82.1} & \textbf{86.2} \\
\method{} (Qwen3.5-122B-A10B) 
& 82.9 & 84.2 & \textbf{85.2} & 84.1 \\
\bottomrule
\end{tabular}
\caption{Reward-modeling comparison on RewardBench2, RM-Bench, and JudgeBench. Unmarked external values are source-reported; ``{*}'' marks our reproduction or re-evaluation, and ``\textdagger'' marks a protocol-limited value. Details are in Appendix~\ref{app:sources-benchmarks}. Bold and underlined numbers mark the highest and second-highest values.}
\label{tab:main}
\vspace{-3mm}
\end{table}

\paragraph{Reward-benchmark protocol.}
Unless otherwise noted, controlled Qwen3.5 LLM-as-a-Judge and Skill-RM runs use each benchmark's original prompt and candidate responses. Skill-RM additionally uses the fixed Reward-Evaluation Skill and its generic resources. Rows marked \emph{+ sample-spec.} mount protocol-specified sample-specific resources when available, such as visible references, sample-specific constraints, or verifier outputs. We report these rows separately because the mounted resources expose per-example evidence beyond the default benchmark protocol.

\paragraph{Baselines and ablations.}
The baselines serve three comparison roles. The main reward-modeling comparison positions \method{} against scalar or learned reward models, including Skywork-Reward-V2~\citep{skyworkrewardv2}; generative or reasoning reward models~\citep{rewardreasoningmodel,rmr1,rationalerm}; rubric-based systems~\citep{autorubric,openrubrics}; and agentic or verifier-based judges~\citep{agenticrewardmodeling,tirjudge}. The Qwen3.5 LLM-as-a-Judge rows provide matched-backbone comparisons that separate the effect of skill-mediated evaluation from model scale or family. Appended-resource prompting, sample-specific resource prompting, and Python-tool access serve as mechanism ablations, exposing additional information or tools without using the Reward-Evaluation Skill.

For controlled Qwen3.5 comparisons, the LLM-as-a-Judge baseline, the \method{} row, the \method{} + sample-spec. row, and mechanism ablations use Qwen3.5-27B~\citep{qwen35blog} with the same benchmark-provided prompt and candidate responses, seed, deterministic decoding settings, and metric denominator within each benchmark. \method{} differs from the LLM-as-a-Judge baseline by giving the judge model access to the Reward-Evaluation Skill; through this interface, the model can inspect applicable resources, invoke enabled checks, record criterion-level evidence, and produce the same benchmark-required output. Qwen3.5-122B-A10B~\citep{qwen35blog} provides an additional open-backbone check. Full evaluation details are provided in Appendix~\ref{app:sources-benchmarks}.

\vspace{-1mm}
\subsection{Main Reward-Modeling Results}
\vspace{-1mm}

Table~\ref{tab:main} reports the main reward-modeling comparison across representative reward models and judge systems.
The central comparison in Table~\ref{tab:main} is the matched Qwen3.5-27B setting. \method{} raises the average from 83.9 to 86.2 relative to the Qwen3.5-27B LLM-as-a-Judge baseline and improves all three benchmarks. Among complete rows, \method{} (Qwen3.5-27B) achieves the strongest average, while \method{} with Qwen3.5-122B-A10B reaches the highest JudgeBench score under a different MoE backbone. We use these cross-family rows for positioning; the next comparison with sample-specific resources and the following ablation focus on matched Qwen3.5-27B settings.

We next report results with sample-specific resources mounted through the Reward-Evaluation Skill. Standard reward benchmarks usually expose only the prompt and candidate responses, so Table~\ref{tab:main} remains the primary comparison. In downstream reward-use scenarios, especially RL, a reward model can also be given references, task constraints, verifier outputs, or other sample-specific evidence. These rows therefore test whether \method{} can use such resources through the Reward-Evaluation Skill rather than treating them as a flat prompt context.

Table~\ref{tab:openrs_resource} shows that sample-specific resources raise the \method{} average from 86.2 to 89.1. OpenRS remains strongest on JudgeBench, consistent with its customized evaluation setup, while \method{} obtains the best average under the same Qwen3.5-27B backbone. This result complements the main comparison in Table~\ref{tab:main}: \method{} improves the matched LLM-as-a-Judge baseline without sample-specific resources, and \method{} + sample-spec. gains further when the protocol provides resources that the judge model can select and aggregate through the skill. The next subsection isolates whether this gain comes from skill-mediated resource organization rather than from resource availability alone.

\begin{table}[t]
\centering
\small
\setlength{\tabcolsep}{4pt}
\renewcommand{\arraystretch}{1.05}
\begin{tabular}{@{}lcccc@{}}
\toprule
\textbf{Method} & \textbf{RewardBench2} & \textbf{RM-Bench} & \textbf{JudgeBench} & \textbf{Avg.} \\
\midrule
Baseline & 81.1 & 89.8 & 80.8 & 83.9 \\
OpenRS & 84.0 & 87.5 & \textbf{93.1} & 88.2 \\
\method{} & 85.0 & \textbf{91.5} & 82.1 & 86.2 \\
\quad + sample-spec. & \textbf{86.0} & \textbf{91.5} & 89.7 & \textbf{89.1} \\
\bottomrule
\end{tabular}
\caption{Reward-modeling comparison with sample-specific resources under the Qwen3.5-27B backbone. ``Sample-spec.'' denotes sample-specific OpenRS evaluation resources mounted.}
\label{tab:openrs_resource}
\vspace{-3mm}
\end{table}
\begin{table}[t]
\centering
\small
\setlength{\tabcolsep}{4pt}
\renewcommand{\arraystretch}{1.05}
\begin{tabular}{@{}lccccc@{}}
\toprule
\textbf{Method} & \textbf{RewardBench2} & \textbf{RM-Bench} & \textbf{JudgeBench} & \textbf{Avg.} & \textbf{$\Delta$} \\
\midrule
Baseline & 81.1 & 89.8 & 80.8 & 83.9 & 0.0 \\
\quad + appended resources & 77.7 & 86.3 & 78.9 & 81.0 & -2.9 \\
\quad + appended + sample-spec. & 84.6 & 74.9 & 86.6 & 82.0 & -1.9 \\
\quad + Python tool & 82.7 & 88.7 & 79.5 & 83.6 & -0.3 \\
\midrule
\method{} & 85.0 & 91.5 & 82.1 & 86.2 & +2.3 \\
\quad + sample-spec. & \textbf{86.0} & \textbf{91.5} & \textbf{89.7} & \textbf{89.1} & \textbf{+5.2} \\
\bottomrule
\end{tabular}
\vspace{-2mm}
\caption{Resource-use ablation with Qwen3.5-27B. ``Sample-spec.'' denotes sample-specific resources, and ``$\Delta$'' is the average-point change from Baseline.}
\label{tab:ablation}
\vspace{-3mm}
\end{table}

\vspace{-1mm}
\subsection{Resource-Use Ablation}
\vspace{-1mm}
Table~\ref{tab:ablation} tests whether the gains come from skill-mediated resource use rather than added context. The upper block keeps the Baseline prompt and directly appends resources, with or without sample-specific resources, or adds Python-tool access. The lower block uses \method{}, then adds the same type of sample-specific resources; sample-spec. abbreviates sample-specific resources.
This same-backbone ablation separates resource availability from resource organization. Directly appended resources lower the average from 83.9 to 81.0, and appending sample-specific resources remains below Baseline at 82.0. Python-tool access is also close to Baseline at 83.6. By contrast, \method{} raises the average to 86.2, and \method{} + sample-spec. reaches 89.1, suggesting that the gains depend on skill-mediated resource organization rather than resource availability or tool access alone.

\vspace{-1mm}
\subsection{Best-of-N Response Selection}
\begin{figure}[t]
\centering
\includegraphics[width=0.73\textwidth]{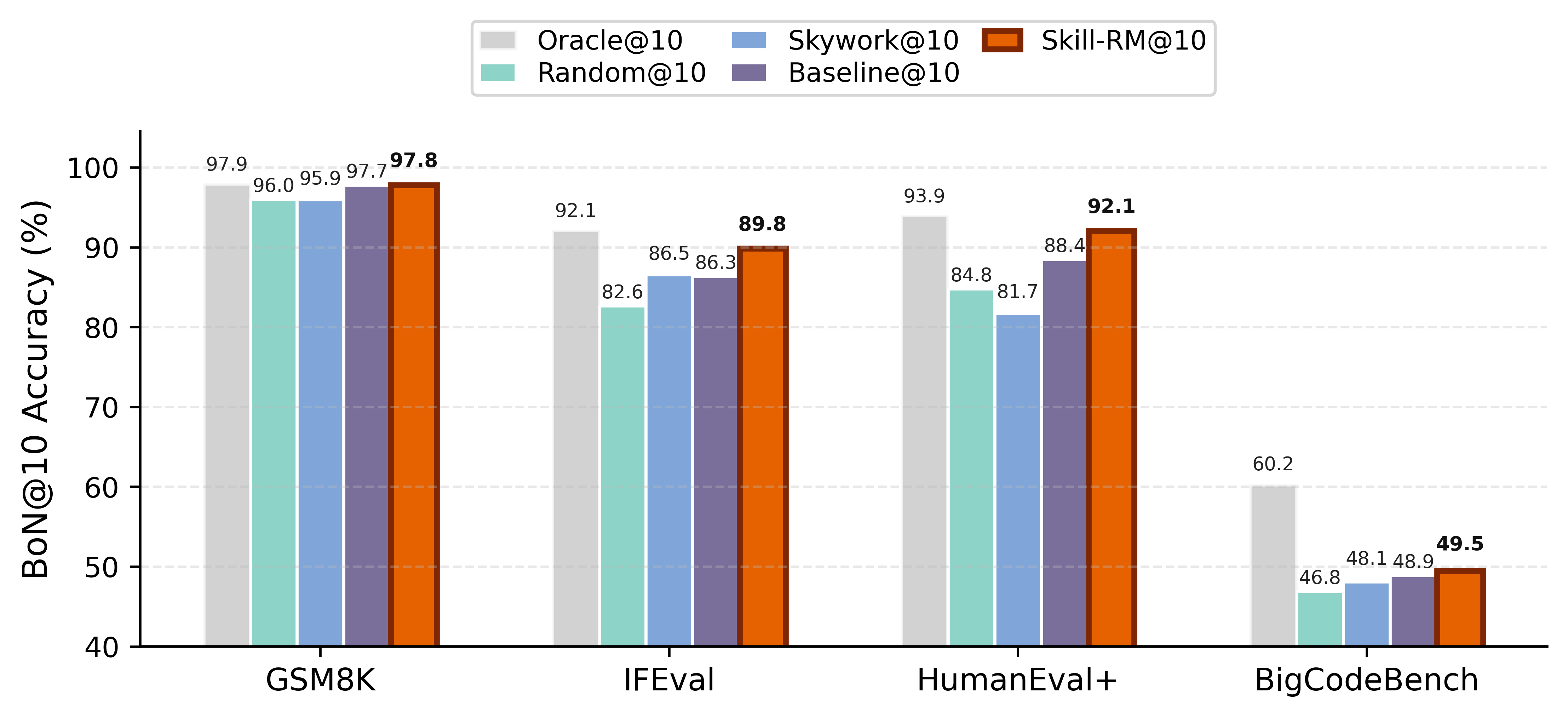}
\vspace{-3mm}
\caption{Best-of-10 response selection accuracy on four JETTS task families. ``Baseline@10'' is Qwen3.5-27B direct judging; ``Oracle@10'' is the pool upper bound.}
\label{fig:bon_at10}
\vspace{-3mm}
\end{figure}

\vspace{-1mm}
Best-of-$N$ response selection uses fixed JETTS reranking pools generated by Qwen2.5-72B-\allowbreak Instruct~\citep{jetts}. Thus, differences reflect selection quality rather than generation. Baseline and \method{} use the same sequential pairwise knockout protocol, while Skywork-Reward-V2-Qwen3-8B scores candidates independently and selects the highest-scored response. Random@10 and Oracle@10 provide random and pool-upper-bound references.
Figure~\ref{fig:bon_at10} gives downstream reward-use evidence for fixed-pool reranking. GSM8K is nearly saturated: \method{} reaches 97.8, close to the Oracle@10 upper bound of 97.9, while Baseline is already 97.7. The clearest gains appear on IFEval and HumanEval+, where \method{} improves over both Baseline and Skywork. BigCodeBench remains harder: \method{} gives a smaller positive gain over Baseline, while Oracle@10 remains much higher. This pattern suggests that skill-mediated reward evaluation is useful for response selection, while code-oriented selection still leaves substantial headroom.

\subsection{Instruction-Following Reward Use}

We further evaluate \method{} in instruction-following reward use, where explicit constraints make reward evidence easier to inspect and the resulting reward can be used for policy optimization. This setting uses an instruction-following Reward-Evaluation Skill that follows the same curation process as the general Reward-Evaluation Skill.
As a judge-side validation before RL, Table~\ref{tab:if_overall} evaluates ranking quality on IF-RewardBench, reporting Kendall correlation on overall assessment for the Single-Turn, Multi-Turn, and System-Prompt subsets. Rows other than Qwen3.5-27B and \method{} use numbers reported by IF-RewardBench; Qwen3.5-27B and \method{} are our runs under the same metric.
\method{} obtains the highest average in Table~\ref{tab:if_overall} (0.524), driven by Single-Turn and Multi-Turn subsets where it reaches 0.619 and 0.540, respectively. Gemini-3-Flash remains stronger on System-Prompt, leaving system-prompt instruction handling as the main gap.

\begin{table}[t]
\centering
\small
\setlength{\tabcolsep}{4pt}
\renewcommand{\arraystretch}{1.05}
\begin{tabular}{@{}lcccc@{}}
\toprule
\textbf{Method} & \textbf{Single-Turn} & \textbf{Multi-Turn} & \textbf{System-Prompt} & \textbf{Avg.} \\
\midrule
\multicolumn{5}{@{}l}{\emph{General LMs}} \\
\addlinespace[1pt]
Gemini-3-Flash & 0.589 & 0.460 & \textbf{0.489} & 0.513 \\
GPT-5-mini & 0.521 & 0.438 & 0.410 & 0.456 \\
Qwen3.5-27B & 0.507 & 0.440 & 0.287 & 0.411 \\
DeepSeek-V3.2 & 0.397 & 0.257 & 0.208 & 0.288 \\
GLM-4.6 & 0.359 & 0.263 & 0.189 & 0.270 \\
Qwen2.5-72B-Instruct & 0.052 & 0.097 & -0.003 & 0.048 \\
\midrule
\multicolumn{5}{@{}l}{\emph{Discriminative RMs}} \\
\addlinespace[1pt]
Skywork-V2-Llama3.1-8B & 0.153 & 0.205 & 0.039 & 0.133 \\
Llama3.1-70B-RM & 0.109 & 0.208 & 0.054 & 0.124 \\
LMUnit-Qwen2.5-72B & 0.126 & 0.108 & 0.009 & 0.081 \\
Qwen2.5-Math-RM & 0.056 & 0.092 & 0.002 & 0.050 \\
\midrule
\multicolumn{5}{@{}l}{\emph{Generative RMs}} \\
\addlinespace[1pt]
RRM-32B & 0.144 & 0.069 & 0.004 & 0.072 \\
RM-R1-Qwen32B & 0.114 & 0.075 & -0.032 & 0.052 \\
\midrule
\multicolumn{5}{@{}l}{\emph{Ours}} \\
\addlinespace[1pt]
\method{} & \textbf{0.619} & \textbf{0.540} & 0.413 & \textbf{0.524} \\
\bottomrule
\end{tabular}
\caption{IF-RewardBench Kendall correlation on overall assessment. Single-Turn, Multi-Turn, and System-Prompt are benchmark subsets; higher is better.}
\label{tab:if_overall}
\end{table}

\begin{table}[t]
\centering
\small
\setlength{\tabcolsep}{4pt}
\renewcommand{\arraystretch}{1.05}
\begin{tabular}{@{}lcccc@{}}
\toprule
\textbf{Method} & \textbf{IFEval} & \textbf{IFBench} & \textbf{AdvancedIF} & \textbf{Avg.} \\
\midrule
Tulu 3 SFT & 72.3 & 21.8 & 17.0 & 37.0 \\
Tulu 3 DPO & 81.0 & 25.9 & 23.9 & 43.6 \\
Tulu 3 & 82.6 & \textbf{27.6} & 25.0 & 45.1 \\
\addlinespace[2pt]
VerIF & 83.7 & \textbf{27.6} & 22.8 & 44.7 \\
\method{} & \textbf{84.8} & \textbf{27.6} & \textbf{25.4} & \textbf{45.9} \\
\bottomrule
\end{tabular}
\caption{Downstream instruction-following RL results. VerIF and Skill-RM use VerInstruct and GRPO-style training; all baseline rows are evaluated from official open-source checkpoints. Scores use prompt-level loose accuracy for IFEval and IFBench, and the macro average over ComplexIF, Carried Context, and System Steerability for AdvancedIF. AdvancedIF is evaluated with our adapted Qwen3.5-122B-A10B judge.}
\label{tab:if_rl}
\end{table}

\paragraph{Downstream RL reward use.}
Table~\ref{tab:if_rl} reports downstream IF-RL results, where all baseline rows are evaluated by us from official open-source checkpoints. VerIF is the closest comparison because both VerIF and Skill-RM use VerInstruct and GRPO-style training; the Tulu rows provide comparisons from different post-training recipes.
\method{} obtains the highest average in Table~\ref{tab:if_rl}, reaching 45.9 compared with 44.7 for VerIF and 45.1 for Tulu 3. Against the closest VerIF comparison, the improvement is clearest on IFEval and AdvancedIF, while IFBench is tied at 27.6. These results support using \method{} as a reward source for instruction-following policy optimization, with modest margins under this evaluation suite. We include a supplementary anchored-pairwise GRPO reward-source ablation in Appendix~\ref{app:downstream-details}.

\vspace{-1mm}
\section{Conclusion}
\vspace{-1mm}


We introduce \textbf{Skill-RM}, a novel framework that reformulates reward modeling as a skill-mediated execution procedure, moving beyond the limitations of monolithic scalar prediction and unstructured prompting. By externalizing evaluation logic into a reusable \textit{Reward-Evaluation Skill}, Skill-RM systematically orchestrates heterogeneous resources and invocation protocols into an explicit computation flow. 
Across reward benchmarks, Skill-RM improves over matched LLM-as-a-Judge baselines under the default benchmark protocol and shows further gains when protocol-specified sample-specific resources are mounted through the skill.
 Downstream studies on fixed-pool best-of-$N$ selection and instruction-following RL further suggest that the same interface can serve as both an offline judge and a reward source. 
Ultimately, this work demonstrates that treating reward evaluation as a structured, reusable, and inspectable procedure is a scalable pathway toward more reliable, transparent, and high-fidelity LLM  alignment.

\paragraph{Limitations and Future Work.}


Despite the effectiveness of Skill-RM, several limitations warrant future investigation. First, our evaluation is currently scoped to text-based instruction-following and standard reward benchmarks. Extending the skill-mediated formulation to multi-modal contexts, long-horizon agentic tasks, or highly open-ended, subjective preference alignment presents a challenging yet promising frontier. Second, the current design relies on manually curated Reward-Evaluation Skills. While this modularity ensures high precision and interpretability, the automation of artifact construction and the mechanisms for continual, self-improving updates remain open problems that could significantly lower the barrier to adoption. Third, the skill-mediated invocation process introduces additional inference overhead compared to traditional single-pass scalar models. Future research on adaptive early stopping, evidence caching, and efficient artifact pruning will be crucial to reconcile the trade-off between evaluation fidelity and computational efficiency.

\bibliography{conference}
\bibliographystyle{conference}

\clearpage
\appendix

\section*{Appendix}

\section{Evaluation Details}
\label{app:sources-benchmarks}

This appendix describes the evaluation protocols needed to interpret the reported results. In Table~\ref{tab:main}, unmarked external rows are source-reported. A superscript \textsuperscript{*} marks a result reproduced or re-evaluated under this paper's protocol. The superscript \textsuperscript{\textdagger} on TIR-Judge-Zero marks the RewardBench2 value reported on five non-Ties subsets; its Avg. value inherits the same mark. The comparison with sample-specific resources is reported in Table~\ref{tab:openrs_resource}; the protocol details are described below.

\paragraph{Benchmark protocols.}
RewardBench2 contains 1,865 examples. We use the best-of-four evaluation compatible with the official protocol, including the separate Ties scoring path, and report the official leaderboard average. RM-Bench contains 11,943 comparisons derived from \(3\times3\) chosen--rejected grids. Source-reported rows use the score reported by the corresponding paper; for our LLM-judge and \method{} runs, each comparison is parsed as win, loss, tie, or error, and ties and errors count as non-wins. JudgeBench contains 620 pairs from the GPT and Claude subsets. Each pair is evaluated twice, with the two responses shown in the original order and then swapped, yielding 1,240 order-level judgments. We normalize both outputs to the same underlying response pair and aggregate them at the pair level: a correct order contributes \(+1\), an incorrect order contributes \(-1\), and a tie or invalid output contributes \(0\); the pair is counted as correct when the sum over the two response orders is positive.

For Best-of-\(N\) response selection, we use the official JETTS response-reranking pools generated by Qwen2.5-72B-Instruct. Each prompt has ten candidate responses, and the main figure reports \(N=10\). The four JETTS task families used in this paper are GSM8K, IFEval, HumanEval+, and BigCodeBench, with public task sizes of 1,319, 541, 270, and 378 examples, respectively. For downstream instruction-following policy evaluation, the IFBench result uses the public IFBench test set with 300 examples, not the multi-turn IFBench package.

\paragraph{Baseline evaluation details.}
For locally marked results, we use each method's native judgment format and convert its outputs to the benchmark metrics used in this paper. For RM-R1-DeepSeek-Distill-Qwen-32B, we reproduce RewardBench2 and JudgeBench: non-Ties RewardBench2 examples use a four-way generative selection prompt, Ties examples use 1--10 candidate ratings passed to the RewardBench2 Ties scorer, and JudgeBench uses the bidirectional pair accuracy described above. RRM-32B is re-evaluated on RewardBench2 with the same generative adapter and RRM-specific parsing for boxed answer and score formats; its RM-Bench and JudgeBench cells remain source-reported.

For OpenRubrics Rubric-RM-8B-voting@5, we follow the released two-stage procedure by generating \(N=5\) rubrics, judging each pair under the generated rubrics, and aggregating by majority vote. Because OpenRubrics is a pairwise judge and does not produce native absolute scores for RewardBench2-Ties, we use a pairwise-to-Ties adapter: all candidate pairs are judged bidirectionally, resolved winners receive one tournament point, unresolved comparisons split half a point to each side, and the resulting normalized scores are passed to the RewardBench2 Ties formula.

For OpenRS in Table~\ref{tab:openrs_resource}, the sample-specific resources are drawn from the public OpenRS evaluation resources\footnote{\url{https://github.com/Qwen-Applications/OpenRS}}. We run the released code with Qwen3.5-27B as the base model and recompute scores under this paper's full-set accuracy-style definitions. The released OpenRS summaries exclude unresolved same or tie outcomes from the denominator; here same, tie, and invalid outputs are counted as non-wins so that OpenRS uses the same denominator convention as the other rows with sample-specific resources. For RewardAgent, we reproduce the no-search setting with Qwen3.5-27B as planner/judger and ArmoRM-Llama3-8B-v0.1 as the scalar reward component. Its RM-Bench and JudgeBench runs use bidirectional pairwise judgments, while its RewardBench2 value skips label-error and Tie/Ties rows and is therefore reported as a sample-level win rate over the remaining examples.

\section{Skill and Prompt Details}
\label{app:skill-prompt-details}

\subsection{Construction of Reward-Evaluation Skills}
\label{app:skill-construction}

The Reward-Evaluation Skills in this work are constructed through an LLM-assisted curation process. This process organizes public or benchmark-permitted evaluation materials into reusable skills; it does not introduce an automatic skill-learning algorithm. Source materials include reward-modeling papers, LLM-as-a-Judge protocols, benchmark documentation, rubrics, references, constraints, checklists, and verifier practices. The curation pipeline uses an LLM-based coding assistant to draft candidate skill instructions and resource entries, while we make the final selection, merging, editing, and rejection decisions.

For each skill, we identify the evaluation criteria, protocol-allowed evidence, and benchmark-required output format. We normalize these materials into resource-bank entries, where each entry records its resource type, content, applicability scope, and access action. The procedural \texttt{SKILL.md} file specifies when the skill is loaded, how resources are selected, what evidence schema is recorded, and how the reward readout maps the structured judgment to the required output. Thus, the final artifact is not a flat prompt containing all materials, but a procedural skill document paired with a resource bank.

We conduct focused manual inspection for source grounding, schema consistency, resource applicability, duplicate entries, and leakage risks. We exclude hidden benchmark labels, direct target preference labels, model identities, answer-order bookkeeping, case-specific solution traces, and feedback from the reported evaluations. We freeze the resulting skills and resource bank before evaluation. Construction-time LLM-assisted interactions are intermediate authoring materials, not part of the inference-time evaluation protocol; the reproducible objects are the frozen skill files, resource indices, invocation protocol, benchmark adapters, and evaluation scripts to be released with the code.

For rows marked \emph{+ sample-spec.}, additional resources come from public OpenRS data and prompt files rather than new annotations in this work. These resources include protocol-exposed answer evidence, instruction-following constraints or checklists, query-type metadata, and the corresponding OpenRS rubric or verification prompts when applicable. They are loaded only in the sample-specific-resource variant and are not folded into the default Reward-Evaluation Skill. OpenRS model identities, answer-order bookkeeping, and target preference labels are not treated as resources available to the judge.

\subsection{Prompt Templates}
Skill-RM keeps the benchmark-provided user prompt and candidate responses unchanged. The system message adds one optional Reward-Evaluation Skill and allows the judge to either answer directly or call \texttt{use\_skill}. The prompt templates used for reward-modeling runs are shown below. The placeholders are filled by the benchmark formatter and the selected Reward-Evaluation Skill.

\subsection{System Prompt of the Reward Judge}
\begin{promptbox}{promptpurple}
\begin{flushleft}
\textbf{Role:}\\
You are an impartial reward judge. Evaluate the candidate responses according to the user's judging request and choose the best available answer. You may optionally load an external judging skill through tool calls. The skill is not loaded by default.

\vspace{0.75em}
\textbf{Benchmark contract:}\\
\{benchmark contract\}: preserve the benchmark's required output format. RewardBench2 uses a best-of-four label \texttt{A/B/C/D}; JudgeBench uses forced-choice \texttt{Output (a)} or \texttt{Output (b)}; RM-Bench uses its scaled pairwise labels.

\vspace{0.75em}
\textbf{Invocation rule:}\\
First judge from the visible prompt and responses. Call \texttt{use\_skill} when exact constraints, code/math checks, factuality, safety/refusal boundaries, or close bias-prone tradeoffs can materially change the verdict. After the call, \texttt{SKILL.md} and the supporting documents become visible. Use at most one skill load, at most \emph{\{max resources\}} viewed resources, and at most \emph{\{max steps\}} assistant turns. When ready, call \texttt{final\_answer} or output the benchmark's required final label.

\vspace{0.75em}
\textbf{Available optional skill:}\\
Available optional skill: \texttt{name} = \emph{\{skill name\}}, \texttt{description} = \emph{\{skill description\}}, \texttt{loading} = \texttt{self-select; SKILL.md and resources are hidden until use\_skill is called}.
\end{flushleft}
\end{promptbox}

\subsection{RewardBench2 User Prompt}
\begin{promptbox}{promptblue}
\begin{flushleft}
Please act as an impartial judge and evaluate the quality of the responses provided by four AI assistants to the user question displayed below. Choose the assistant that follows the user's instructions and answers the user's question best. Consider helpfulness, relevance, accuracy, depth, creativity, and level of detail. Avoid position, length, and assistant-name biases. After a short explanation, output exactly \texttt{[[A]]}, \texttt{[[B]]}, \texttt{[[C]]}, or \texttt{[[D]]}.

\vspace{0.75em}
\textbf{[User Question]}\\
\emph{\{question\}}

\vspace{0.75em}
\textbf{[The Start of Assistant A's Answer]}\\
\emph{\{answer\_a\}}\\
\textbf{[The End of Assistant A's Answer]}

\vspace{0.75em}
\textbf{[The Start of Assistant B's Answer]}\\
\emph{\{answer\_b\}}\\
\textbf{[The End of Assistant B's Answer]}

\vspace{0.75em}
\textbf{[The Start of Assistant C's Answer]}\\
\emph{\{answer\_c\}}\\
\textbf{[The End of Assistant C's Answer]}

\vspace{0.75em}
\textbf{[The Start of Assistant D's Answer]}\\
\emph{\{answer\_d\}}\\
\textbf{[The End of Assistant D's Answer]}.
\end{flushleft}
\end{promptbox}

\subsection{Tool Interface}
\begin{promptbox}{promptgreen}
\begin{flushleft}
\textbf{Before skill loading:}\\
The available tools are \texttt{use\_skill} and \texttt{final\_answer}.

\vspace{0.75em}
\textbf{After skill loading:}\\
The judge can additionally call \texttt{view\_resource}, \texttt{python\_sandbox}, and enabled verifier resources.

\vspace{0.75em}
\textbf{Sandbox scope:}\\
The Python sandbox receives only the visible prompt and candidate responses.
\end{flushleft}
\end{promptbox}

\subsection{Loaded Reward-Evaluation Skill}

The core skill loaded in reward-modeling runs without sample-specific resources is shown below. It is the \texttt{SKILL.md} document for \texttt{reward\_judge\_fair}; supporting resources remain hidden until the judge loads the skill and are summarized separately in Table~\ref{tab:supporting_resource_index}.

\begin{skillbox}
\begin{flushleft}
\textbf{Metadata}\\
name: reward\_judge\_fair\\
description: Use this Skill-RM reward judge to compare candidate responses for a visible user request with generic rubric, principles, bias controls, output contract, and Python sandbox checks over visible text.\\
metadata:\\
\hspace*{1em}family: reward\_judge\\
\hspace*{1em}short-description: Generic visible-text reward judge\\
\hspace*{1em}method: skill\_fair

\vspace{0.75em}
\textbf{Reward Judge}\\
Use this skill to organize a reward judgment from the current user request and candidate responses. The skill is a controller and resource interface, not a per-sample prompt template.

\vspace{0.75em}
\textbf{Inputs}\\
The host message provides only:\\
\hspace*{1em}- the visible user prompt or instruction;\\
\hspace*{1em}- candidate responses and their current labels;\\
\hspace*{1em}- the required final output format.\\
Use the current prompt and candidate responses as the full task context.

\vspace{0.75em}
\textbf{Resource Interface}\\
After this skill is loaded, use only resources listed in the current resource index. The resources are generic:\\
\hspace*{1em}- rubric: generic reward judging criteria;\\
\hspace*{1em}- principle: generic correctness, instruction-following, safety, usefulness, and anti-style-bias principles;\\
\hspace*{1em}- calibration: position, verbosity, style, and confidence-bias controls;\\
\hspace*{1em}- aggregation: generic evidence-combination policy;\\
\hspace*{1em}- output\_contract: JSON verdict contract;\\
\hspace*{1em}- tool: python\_sandbox, which can inspect only the visible prompt and candidate responses.

\vspace{0.75em}
\textbf{Tool Use}\\
Use view\_resource to read generic rubric, principles, bias control, aggregation, or output format resources.\\
Use python\_sandbox when deterministic checking over visible text can change the verdict. It runs short Python over only:\\
\hspace*{1em}- prompt: the visible user prompt;\\
\hspace*{1em}- candidates: the current visible candidate responses keyed by label;\\
\hspace*{1em}- sample: \{"prompt": prompt, "candidates": candidates\}.\\
Use it for counts, regex/format checks, JSON/list structure, simple arithmetic, supplied examples, small code-behavior checks, or answer extraction from visible candidate text.\\
run\_resource should normally not be used with this skill. Read generic resources with view\_resource, use python\_sandbox for deterministic visible-text checks, then submit final\_answer.

\vspace{0.75em}
\textbf{Decision Procedure}\\
1. Identify the user's actual task and mandatory constraints from the prompt.\\
2. Compare candidates under one shared criterion.\\
3. Prioritize hard correctness, instruction following, safety, factuality, and required output format.\\
4. Use python\_sandbox only for checks that can be computed from visible prompt/candidates.\\
5. Apply bias controls: do not prefer position, length, markdown polish, confidence, or fluent style unless it improves task success.\\
6. Use Tie only when candidates are genuinely equivalent or the visible evidence is insufficient for a reliable preference.\\
7. Return the required JSON.

\vspace{0.75em}
\textbf{Output}\\
Return JSON only:\\
\{\\
\hspace*{1em}"verdict": "A|B|Tie",\\
\hspace*{1em}"confidence": 0.0,\\
\hspace*{1em}"used\_resources": [],\\
\hspace*{1em}"reason": "short reason"\\
\}
\end{flushleft}
\end{skillbox}

\Needspace{0.48\textheight}
\subsection{Supporting Resource Index}

The \texttt{reward\_judge\_fair} skill contains generic resources. The \texttt{reward\_judge\_operational} skill keeps the same core judging style and additionally exposes benchmark-level rubrics and sample-specific resources when present.

\begin{table}[H]
\centering
\small
\setlength{\tabcolsep}{4pt}
\renewcommand{\arraystretch}{1.08}
\begin{tabularx}{\textwidth}{@{}>{\raggedright\arraybackslash}p{0.22\textwidth}>{\raggedright\arraybackslash}p{0.35\textwidth}Y@{}}
\toprule
\textbf{Resource} & \textbf{Content} & \textbf{Use} \\
\midrule
\texttt{rubric} & Generic reward-judging criteria. & Defines the shared criteria used to compare candidate responses. \\
\texttt{principle} & Correctness, instruction-following, safety, usefulness, and anti-style-bias principles. & Prioritizes hard task success over superficial style or confidence. \\
\texttt{calibration} & Position, verbosity, style, and confidence-bias controls. & Reduces judge sensitivity to presentation artifacts that do not affect answer quality. \\
\texttt{aggregation} & Generic evidence-combination policy. & Specifies how criterion-level evidence is combined into a final preference or tie. \\
\texttt{output\_contract} & JSON verdict contract. & Constrains the loaded skill output before benchmark-specific parsing. \\
\texttt{python\_sandbox} & A deterministic checker over the visible prompt and candidate responses. & Supports counts, regular expressions, arithmetic, format checks, supplied examples, and answer extraction. \\
\texttt{sample-specific resources} & OpenRS-derived task fields, answer evidence, and constraint evidence, available only for rows marked \emph{+ sample-spec.}. & Provides per-instance references or constraints after labels, model identities, answer-order bookkeeping, and direct target labels are removed. \\
\bottomrule
\end{tabularx}
\caption{Supporting resource index for the Reward-Evaluation Skill. The first six rows are generic resources; the final row describes the sample-specific-resource extension.}
\label{tab:supporting_resource_index}
\end{table}

\paragraph{Ablation implementations.}
The appended-resource ablation removes skill selection and progressive disclosure. It places the skill document and resource documents directly before the benchmark judging prompt. In the sample-specific variant, the same OpenRS per-sample documents are appended as ordinary prompt text. The Python-tool ablation keeps the baseline benchmark prompt but adds \texttt{python\_sandbox} and \texttt{final\_answer}, without providing the skill document or resource documents.

\paragraph{Inference settings.}
We serve the Qwen3.5 judge backbones with vLLM\footnote{\url{https://github.com/vllm-project/vllm}} 0.17.1 on 8 NVIDIA A800 GPUs and query them through vLLM's OpenAI-compatible API. Judge calls use deterministic decoding unless otherwise stated: temperature \(0\), top-\(p=1.0\), maximum generation length 4096, and thinking disabled. Tool-enabled runs use vLLM tool calling with a Qwen-compatible parser. The same inference settings are used for the LLM-as-a-Judge baseline, appended-resource ablations, and Skill-RM runs.

\section{Additional Backbone Results}
\label{app:additional-results}

Table~\ref{tab:additional_multibackbone_results} checks whether the effect of \method{} is tied to a single judge backbone. Each block compares the direct LLM-as-a-Judge baseline with \method{} without sample-specific resources under the same Qwen3.5 backbone. The + sample-spec. row additionally provides protocol-exposed sample-specific resources, when available, to the skill.

\begin{table}[htbp]
\centering
\small
\setlength{\tabcolsep}{4pt}
\renewcommand{\arraystretch}{1.05}
\begin{tabular}{@{}lcccc@{}}
\toprule
\textbf{Method} & \textbf{RewardBench2} & \textbf{RM-Bench} & \textbf{JudgeBench} & \textbf{Avg.} \\
\midrule
\multicolumn{5}{@{}l}{\emph{Qwen3.5-9B}} \\
\addlinespace[1pt]
LLM-as-a-Judge & 61.0 & 64.1 & 57.3 & 60.8 \\
\method{} & 62.9 & \textbf{73.0} & \textbf{62.7} & \textbf{66.2} \\
\quad + sample-spec. & \textbf{63.0} & 72.8 & 61.3 & 65.7 \\
\midrule
\multicolumn{5}{@{}l}{\emph{Qwen3.5-27B}} \\
\addlinespace[1pt]
LLM-as-a-Judge & 81.1 & 89.8 & 80.8 & 83.9 \\
\method{} & 85.0 & \textbf{91.5} & 82.1 & 86.2 \\
\quad + sample-spec. & \textbf{86.0} & \textbf{91.5} & \textbf{89.7} & \textbf{89.1} \\
\midrule
\multicolumn{5}{@{}l}{\emph{Qwen3.5-35B-A3B}} \\
\addlinespace[1pt]
LLM-as-a-Judge & 78.2 & 87.5 & 75.0 & 80.2 \\
\method{} & 81.1 & \textbf{88.6} & \textbf{76.9} & 82.2 \\
\quad + sample-spec. & \textbf{84.4} & 88.3 & 76.5 & \textbf{83.0} \\
\midrule
\multicolumn{5}{@{}l}{\emph{Qwen3.5-122B-A10B}} \\
\addlinespace[1pt]
LLM-as-a-Judge & 79.0 & 82.2 & 67.1 & 76.1 \\
\method{} & 82.9 & \textbf{84.2} & 85.2 & 84.1 \\
\quad + sample-spec. & \textbf{84.8} & 81.9 & \textbf{91.5} & \textbf{86.0} \\
\bottomrule
\end{tabular}
\caption{Additional multi-backbone results on RewardBench2, RM-Bench, and JudgeBench. Sample-spec. denotes sample-specific resources. Values are percentages, and bold marks the best value within each backbone block.}
\label{tab:additional_multibackbone_results}
\end{table}

Across all four backbones, \method{} without sample-specific resources improves the average over the corresponding LLM-as-a-Judge baseline, suggesting that the skill formulation is not tied to a single judge model. The trend is not monotonic with model size: the dense 27B backbone is stronger than the 122B-A10B LLM-as-a-Judge baseline on this suite, while the 122B-A10B model benefits more from skill-mediated reward evaluation than from direct prompting. Sample-specific resources also show a boundary: they improve the average for the 27B, 35B-A3B, and 122B-A10B backbones, but not for the 9B model. This pattern suggests that additional resources are not uniformly beneficial; smaller backbones may be less reliable at selecting and applying them. We therefore treat Table~\ref{tab:additional_multibackbone_results} as robustness and boundary evidence, while the same-backbone ablation in Table~\ref{tab:ablation} remains the main mechanism test.

\section{Instruction-Following RL Details}
\label{app:downstream-details}

This section documents the reward protocol behind the downstream instruction-following results in Table~\ref{tab:if_rl}. The main run uses a pointwise reward: each rollout response is judged against the visible instruction and mapped to a scalar score in \([0,1]\). The policy starts from Llama-3.1-Tulu-3-8B-SFT and is trained on VerInstruct-derived prompts with GRPO in Verl\footnote{\url{https://github.com/volcengine/verl}}. The judge model is Qwen3.5-27B with deterministic decoding.

\begin{table}[H]
\centering
\small
\setlength{\tabcolsep}{5pt}
\renewcommand{\arraystretch}{1.08}
\begin{tabularx}{0.80\textwidth}{@{}lX@{}}
\toprule
\textbf{Setting} & \textbf{Value} \\
\midrule
Policy initialization & Llama-3.1-Tulu-3-8B-SFT \\
Training data & VerInstruct \\
RL trainer & Verl 0.8.0.dev with GRPO \\
Python version & Python 3.12 \\
Prompt/response length & 1024 / 2048 tokens \\
Train batch size & 32 prompts \\
Rollouts per prompt & 4 \\
PPO mini-batch size & 32 \\
PPO micro-batch size & 4 per GPU \\
Actor learning rate & \(10^{-6}\) \\
KL loss coefficient & 0.001 \\
Rollout sampling & Temperature \(1.0\), top-\(p=1.0\) \\
Reward judge & Qwen3.5-27B, temperature \(0\), top-\(p=1.0\), max tokens 4096 \\
Reward skill & Instruction-following Reward-Evaluation Skill with VerInstruct verifier resources, static instruction-following resources, and Python checks \\
Training duration & 1 epoch on 8 A800 training GPUs \\
Reward service & 16 A800 GPUs for Skill-RM inference \\
\bottomrule
\end{tabularx}
\caption{Training settings for the main instruction-following RL experiment.}
\label{tab:if_rl_training_settings}
\end{table}

\paragraph{Pointwise reward.}
The instruction-following Reward-Evaluation Skill decomposes the instruction into atomic constraints, checks hard requirements before style or preference, and returns three fields: \texttt{satisfied\_count}, \texttt{total\_count}, and \texttt{score}. When a VerInstruct checklist is available, the satisfied-to-total ratio provides the base reward. When checker code is available, the judge executes protocol-visible checker code or Python checks over the prompt, response, system prompt, and dialogue history. Invalid judge outputs or failed reward calls receive a fallback reward of \(0.5\), so parser failures do not become either positive or negative training labels.

\paragraph{Anchored-pairwise ablation.}
Table~\ref{tab:if_rl_ablation} reports a supplementary anchored-pairwise GRPO ablation. It asks whether a pairwise Skill-RM judgment can be converted into a scalar reward by comparing each rollout with a fixed reference response generated by the initial Tulu-3 SFT policy. During training, the rollout side is randomized to reduce position bias. A rollout win maps to reward \(1\), a loss maps to reward \(0\), and a failed or unresolved judgment maps to reward \(0.5\). The checklist-only and verification-only rows use separate resource subsets; the main pointwise reward in Table~\ref{tab:if_rl} uses the full instruction-following Reward-Evaluation Skill.

\begin{table}[H]
\centering
\small
\setlength{\tabcolsep}{4pt}
\renewcommand{\arraystretch}{1.05}
\begin{tabular}{@{}lcccc@{}}
\toprule
\textbf{Reward variant} & \textbf{IFEval} & \textbf{IFBench} & \textbf{AdvancedIF} & \textbf{Avg.} \\
\midrule
\multicolumn{5}{@{}l}{\emph{Direct and tool-only}} \\
\addlinespace[1pt]
Qwen3.5-27B LLM-as-a-Judge & 75.8 & 24.8 & 21.4 & 40.7 \\
Qwen3.5-27B + Python tool & 81.0 & 25.5 & 22.8 & 43.1 \\
\midrule
\multicolumn{5}{@{}l}{\emph{\method{} without verifier resources}} \\
\addlinespace[1pt]
\method{} checklist & 82.8 & 26.5 & 25.1 & 44.8 \\
\method{} verification & 82.1 & 25.9 & 25.8 & 44.6 \\
\midrule
\multicolumn{5}{@{}l}{\emph{\method{} with verifier fields in prompt}} \\
\addlinespace[1pt]
\method{} checklist & \textbf{84.7} & 26.2 & \textbf{26.0} & \textbf{45.6} \\
\method{} verification & 83.9 & 26.5 & 24.0 & 44.8 \\
\midrule
\multicolumn{5}{@{}l}{\emph{\method{} with verifier fields as resources}} \\
\addlinespace[1pt]
\method{} checklist & 82.1 & \textbf{27.2} & 23.9 & 44.4 \\
\method{} verification & 83.2 & 25.9 & 24.7 & 44.6 \\
\bottomrule
\end{tabular}
\caption{Supplementary instruction-following RL reward-variant ablation under an anchored pairwise GRPO protocol. Verifier fields denote the VerInstruct checker/function fields associated with each instance; metrics follow Table~\ref{tab:if_rl}.}
\label{tab:if_rl_ablation}
\end{table}

The anchored-pairwise ablation supports the same qualitative trend as the main result: Skill-RM-style rewards outperform direct or tool-only judging, and the presentation of verification evidence also matters. Adding only a Python tool improves the LLM-as-a-Judge average from 40.7 to 43.1, while \method{} rewards reach higher averages across all variants. Prompted verifier fields give the best average, driven by IFEval and AdvancedIF, whereas exposing the same fields through the skill gives the best IFBench score. Downstream RL is therefore sensitive not only to which verification evidence is available, but also to how the judge model is asked to use it.

\end{document}